\pdfoutput=1

\documentclass[11pt]{article}

\usepackage[]{acl}

\usepackage{times}
\usepackage{latexsym}

\usepackage[T1]{fontenc}

\usepackage[utf8]{inputenc}

\usepackage{microtype}
\usepackage{mathrsfs}
\usepackage{amsmath}
\usepackage{amssymb}
\usepackage{graphicx}
\usepackage{algorithm}
\usepackage{algpseudocode}
\usepackage{multirow}
\usepackage{booktabs}
\usepackage{url}
\usepackage{lipsum}
\usepackage{enumitem}
\usepackage{subfigure}

\title{Can Question Rewriting Help Conversational Question Answering?}

\author{Etsuko Ishii\thanks{\hspace{2mm}Equal Contribution},\ Yan Xu$^*$, Samuel Cahyawijaya$^*$, Bryan Wilie\\
  The Hong Kong University of Science and Technology\\
  \texttt{\{eishii, yxucb, scahyawijaya, bwilie\}@connect.ust.hk} \\
  }

\begin{document}
\maketitle
\begin{abstract}
Question rewriting (QR) is a subtask of conversational question answering (CQA) aiming to ease the challenges of understanding dependencies among dialogue history by reformulating questions in a self-contained form. Despite seeming plausible, little evidence is available to justify QR as a mitigation method for CQA. 
To verify the effectiveness of QR in CQA, we investigate a reinforcement learning approach that integrates QR and CQA tasks and does not require corresponding QR datasets for targeted CQA.
We find, however, that the RL method is on par with the end-to-end baseline. We provide an analysis of the failure and describe the difficulty of exploiting QR for CQA.
\end{abstract}

\section{Introduction}
The question rewriting (QR) task has been introduced as a mitigation method for conversational question answering (CQA). CQA asks a machine to answer a question based on the provided passage and a multi-turn dialogue~\citep{reddy-etal-2019-coqa, choi-etal-2018-quac}, which poses an additional challenge to comprehend the dialogue history. To ease the challenge, QR aims to teach a model to paraphrase a question into a self-contained format using its dialogue history~\citep{elgohary-etal-2019-unpack, anantha-etal-2021-open}.
Except for \citet{kim-etal-2021-learn}, however, no one has provided evidence that QR is effective for CQA in practice. Existing works on QR often (i) depend on the existence of a QR dataset for every target CQA dataset, and (ii) focus more on generating high-quality rewrites than improving CQA performance, making them unsatisfactory for the justification of QR.

To verify the effectiveness of QR, we explore a reinforcement learning (RL) approach that integrates QR and CQA tasks without corresponding labeled QR datasets. In the RL framework, a QR model plays the role of ``the agent'' that receives rewards from a QA model that acts as ``the environment.'' During training, the QR model aims to maximize the performance on the CQA task by generating better rewrites of the questions.

\begin{figure}[!t]
	\centering
	\includegraphics[width=0.9\linewidth]{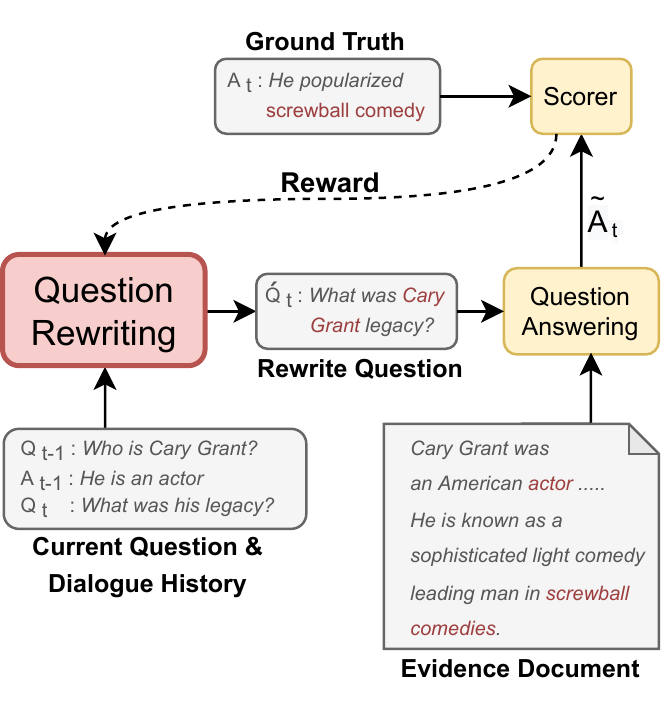}
	\caption{Overview of the RL approach. The current question $Q_t$ and its dialogue history are reformulated into a self-contained question $\acute{Q}_t$ by the QR model. Then, $\acute{Q}_t$ is passed to the QA model to extract an answer span $\Tilde{A}_t$ from the evidence document. We train the QR model by maximizing the reward obtained by comparing the predicted answer span $\Tilde{A}_t$ with the gold span $A_t$.}
	\vspace{-10pt}
	\label{fig:method-comparison}
\end{figure}

Despite the potential and plausibility of the RL approach, our experimental results suggest an upper bound of the performance, and it is on par with the baselines without QR. 
In this paper, we provide analysis to (i) understand the reason for the failure of the RL approach and (ii) reveal that QR cannot improve CQA performance even with the non-RL approaches. The code is available at \url{https://github.com/HLTCHKUST/cqr4cqa}.

\begin{table*}[!ht]
\centering
\resizebox{\textwidth}{!}{
    \begin{tabular}{llccccccccc}
    \toprule
    \multicolumn{2}{l}{\multirow{2}{*}{\textbf{Models}}} & \multicolumn{6}{c}{\textbf{CoQA}} & \multicolumn{3}{c}{\textbf{QuAC}} \\ \cmidrule(lr){3-8} \cmidrule(lr){9-11}
    \multicolumn{2}{l}{} & \textbf{Overall F1} & \textbf{Child.} & \textbf{Liter.} & \textbf{M\&H} & \textbf{News} & \textbf{Wiki.} & \textbf{F1} & \textbf{HEQ-Q} & \textbf{HEQ-D} \\ \midrule
    & end-to-end & 84.5 & \textbf{84.4} & 82.4 & \textbf{82.9} & 86.0 & 86.9 & \textbf{67.8} & 63.5 & 7.9 \\ \midrule
     \multirow{2}{*}{QReCC} 
    & pipeline & 82.9 & 82.9 & 80.9 & 81.5 & 84.4 & 84.8 & 66.3 & 62.0 & 6.6 \\ 
    & ours & \underline{\textbf{84.7}} & \underline{84.3} & \underline{\textbf{83.1}} & \underline{82.7} & \underline{\textbf{86.3}} & \underline{86.8} & \underline{67.6} & \underline{63.2} & \underline{7.8} \\ \midrule
    \multirow{3}{*}{CANARD} 
    & pipeline & 82.8 & 83.4 & 80.1 & 80.8 & 84.4 & 85.6 & 66.5 & 62.5 & 7.4 \\
    & EXCORD$^\dagger$ & 83.4 (+0.6) & \underline{\textbf{84.4 (1.9)}} & 81.2 (+1.0) & 79.8 (-0.3) & 84.6 (+0.3) & \underline{\textbf{87 (0.0)}} & \underline{67.7 (+1.2)} & \underline{\textbf{64.0 (+1.6)}} & \underline{\textbf{9.3 (+2.1)}} \\ 
    & ours & \underline{84.4} & 84.1 & \underline{82.7} & \underline{82.6} & \underline{86.0} & 86.7 & 67.4 & 62.7 & 8.1  \\ \bottomrule 
    \end{tabular}
}
\caption{Evaluation results of our approach and baselines on the test set. EXCORD$^\dagger$ follows the results reported by~\citet{kim-etal-2021-learn} and ($\pm$x.x) indicate the improvement compared to their original baseline. \textbf{Bold} are the best results amongst all. \underline{Underlined} represents the best score on each combination of the CQA and QR datasets. }
\label{tab:results}
\end{table*}

\section{Related Work}
The CQA task aims to assist users in seeking information~\citep{reddy-etal-2019-coqa,choi-etal-2018-quac,campos-etal-2020-doqa}. The key challenge is to resolve the conversation history and understand a highly-contextualized question. Most prior works focus on model structures~\citep{zhu2018sdnet,yeh2019flowdelta,zhang2021retrospective,zhao2021ror}
or training techniques~\cite{ju2019technical,xu2021caire} to improve the performance. 
QR tasks have been proposed to further improve CQA systems by paraphrase a question into a self-contained styles~\citep{elgohary-etal-2019-unpack, hansen2020what, anantha-etal-2021-open}.
While many of the existing works on QR put more effort toward generating high-quality rewrites~\citep{lin2020conversational, vakulenko2021question}, \citet{kim-etal-2021-learn} introduced a framework to leverage QR to finetune CQA models with a consistency-based regularization.
QR has also been studied in single-turn QA and other information-seeking tasks~\citep{nogueira-cho-2017-task, buck2018ask}. 

\section{Methodology}
We denote a CQA dataset as $\{\mathcal{D}^n\}_{n=1}^N$ and the dialogue history at turn $t$ as $\mathcal{D}_{t} = \{ (Q_i, A_i) \}_{i=1}^t$, where $Q_t$ is the question and $A_t$ is the answer. Along with the QA pairs, the corresponding evidence documents $Y_t$ are also given.

As depicted in Figure~\ref{fig:method-comparison}, our proposed RL framework involves a QA model as an environment and a QR model as an agent. 
Let $\acute{Q}_t = \{\acute{q}_l \}_{l=1}^L$ denote a generated rewritten question sequence of $Q_t$. The objective of the QR model is to rewrite the question $Q_t$ at turn $t$ into a self-contained version, based on the current question and the dialogue history $\mathcal{D}_{t-1}$. 
The agent takes an input state $X_t = (\mathcal{D}_{t-1}, Q_t)$ and generates a paraphrase $\acute{Q}_t$. Then, $\acute{X}_t = (\mathcal{D}_{t-1}, \acute{Q}_t)$ and an evidence document $Y_t$ are provided to an environment, namely, the QA model $f_{\phi}$, which extracts an answer span $\Tilde{A}_t = f_{\phi}(\acute{X_t}, Y_t)$.
We aim for the agent, a QR model $\pi_\theta$, to learn to generate a high-quality paraphrase of the given question based on the reward received from the environment.

The policy, in our case the QR model, assigns probability
\begin{align}
    \pi_\theta (\acute{Q}_t | X_t) = \prod_{l=1}^L p (\acute{q}_l | \acute{q}_1, \ldots, \acute{q}_{l-1}, X_t).
\end{align}
Our goal is to maximize the expected reward of the answer returned under the policy, namely,
\begin{align}
    \mathrm{E}_{\acute{q}_t \sim \pi_\theta(\cdot|q_t)} [r (f_{\phi}(\acute{X_t}))],
\label{eq:expected_reward}
\end{align}
where $r$ is a reward function. We apply the token-level F1-score between the predicted answer span $\Tilde{A}_t$ and the gold span $A_t$ as the reward $r$. We can directly optimize the expected reward in Eq.~\ref{eq:expected_reward} using RL algorithms. 

Prior to the training process, the QA model $f_{\phi}$ is fine-tuned on $\{\mathcal{D}^{n}\}$ and the QR model is initialized with $\pi_\theta = \pi_{\theta_0}$, where $\pi_{\theta_0}$ is a pretrained language model. We apply Proximal Policy Optimization (PPO) \citep{schulman2017ppo, ziegler2019finetuning} to train $\pi_\theta$. PPO is a policy gradient method which alternates between sampling data through interaction with the environment and optimizing a surrogate objective function via stochastic gradient ascent. 
Following~\citet{ziegler2019finetuning}, we apply a KL-penalty to the reward $r$ so as to prevent the policy $\pi_\theta$ from drifting too far away from $\pi_{\theta_0}$:
\begin{align*}
    R_t = R(\acute{X}_t) = r(f_{\phi}(\acute{X}_t)) - \beta \mathrm{KL} (\pi_\theta, \pi_{\theta_0}),
\end{align*}
where $\beta$ represents a weight factor and $R_t$ is the modified reward of $r$.

\begin{table*}[t]
\centering
\resizebox{\linewidth}{!}{
\begin{tabular}{llc|llc}
\toprule
\multicolumn{2}{c}{Question}                                                                                                                                                                                                    & \multicolumn{1}{c|}{F1 Score}                                                                & \multicolumn{2}{c}{Question}                                                                                                                                           & \multicolumn{1}{c}{F1 Score}                                                              \\ \midrule
$Q_t$                     & What is the Vat the {\color[HTML]{9A0000} \textbf{library}} of?                                    & 1.0                          & $Q_t$                      & Where {\color[HTML]{9A0000} \textbf{did}} the band The Smashing Pumpkins put on display?                & 1.0                         \\
$\acute{Q}_t$                    &    What is the Vat the {\color[HTML]{9A0000} \textbf{Library}} of? & 0.22   & $\acute{Q}_t$                      & Where {\color[HTML]{9A0000} \textbf{was}} the band The Smashing Pumpkins put on display? & 0.0                         \\
\midrule
$Q_t$    & What was {\color[HTML]{9A0000} \textbf{everybody}} doing?                & 0.91  & $Q_t$  &  Which company produced the movie {\color[HTML]{9A0000}\textbf{ Island of Misfit Toys}}?                                     & 1.0 \\
$\acute{Q}_t$        & What was {\color[HTML]{9A0000} \textbf{everyone}} doing?   & 0.0   & $\acute{Q}_t$    &   Which company produced the movie{\color[HTML]{9A0000}\textbf{, The Island of Misfit Toys}}? & 0.0                         \\
\bottomrule
\end{tabular}
}
\caption{Minor modification of questions may cause a drastic change in CQA performance.}
\label{tab:case_study}
\end{table*}

\section{Experiments}
\subsection{Setup}
We use a pretrained RoBERTa~\cite{liu2019roberta} model as the initial QA model and adapt it to the CQA tasks. 
For the QR models, we leverage pretrained GPT-2~\cite{radford2019language} and first finetune them with QR datasets for better initialization.
We attempt three settings: (a) directly fine-tune the QA model on the CQA datasets (end-to-end), (b) fine-tune the QA model with questions rewritten by the QR model (pipeline), and (c) train the QR model based on the reward obtained from the QA model. 
More details of the experiments can be found in Appendix~\ref{app:main_experiments}.

\paragraph{Datasets}
We conduct our experiments on two crowd-sourced CQA datasets, CoQA~\citep{reddy-etal-2019-coqa} and QuAC~\citep{choi-etal-2018-quac}.
Since the test set is not publicly available for both CoQA and QuAC, following \citet{kim-etal-2021-learn}, we randomly sample 5\% of dialogues in the training set and adopt them as our validation set and report the test results on the original development set for the CoQA experiments. We apply the same split as \citet{kim-etal-2021-learn} for the QuAC experiments.

For the QR model pre-training, we use two QR datasets: QReCC~\cite{anantha2021open} and CANARD~\cite{elgohary2019can}. CANARD is generated by rewriting a subset of the original questions in the QuAC datasets, and contains 40K questions in total. QReCC is built upon three publicly available datasets: QuAC, TREC Conversational Assistant Track (CAsT)~\cite{dalton2020trec} and Natural Questions (NQ)~\cite{kwiatkowski2019natural}. QReCC contains 14K dialogues with 80K questions, and 9.3K dialogues are from QuAC.



\paragraph{Evaluation Metrics}
Following the leaderboards, we utilize the unigram F1 score to evaluate the QA performance.
In CoQA evaluation, the QA models are also evaluated with the domain-wise F1 score.
In QuAC evaluation, we incorporate the human equivalence score HEQ-Q and HEQ-D as well. HEQ-Q indicates the percentage of questions on which the model outperforms human beings and HEQ-D represents the percentage of dialogues on which the model outperforms human beings for all questions in the dialogue.

\subsection{Results}
We report our experimental results in Table~\ref{tab:results}. We see that our RL approach yields 0.9--1.6 F1 improvement over the pipeline setting regardless of the dataset combinations and performs almost as well as the end-to-end setting.
This partially supports our expectation that RL lifts the CQA performance. However, we find it almost impossible to bring significant improvement over the end-to-end baseline despite our extensive trials.
One reason why we cannot provide as much improvement as reported in~\citet{kim-etal-2021-learn} would be related to the inputs of the QA model. Their EXCORD feeds the original questions together with the rewritten questions, whereas we only use the rewritten questions. It is also noteworthy that their results are consistently lower than ours, even lower than our end-to-end settings.

Our inspection of the questions generated by the QR models reveals that the models learn to copy the original questions by PPO training, and this is the direct reason that our method cannot outperform the end-to-end baselines. Indeed, on average, 89.6\% of the questions are the same as the original questions after PPO training, although this value is 34.5\% in the pipeline settings. 
We also discover a significant correlation between the performance and how much the QR models copy the original question (the correlation coefficient is 0.984 for CoQA and 0.967 for QuAC) 
and the edit distance from the original question (the correlation coefficient is -0.996 for CoQA and -0.989 for QuAC).

\begin{table}[!t]
\centering
\resizebox{\linewidth}{!}{%
\begin{tabular}{lllll}
\toprule
\multirow{2}{*}{\textbf{Perturb}} & \multicolumn{2}{c}{\textbf{Sentiment Analysis}} &  \multicolumn{2}{c}{\textbf{CQA}} \\  
\cmidrule(lr){2-3} \cmidrule(lr){4-5}
 & \multicolumn{1}{c}{\textbf{Amazon}} & \multicolumn{1}{c}{\textbf{Yelp}} & \multicolumn{1}{c}{\textbf{CoQA}} & \multicolumn{1}{c}{\textbf{QuAC}} \\ 
\midrule
\textbf{Original}  & 95.8  & 98.2  & 84.5  & 67.8 \\
\textbf{UPC}  & 95.8 (-)  & 96.7  {\color[HTML]{00008B}\textbf{(-1.5)}}  & 74.8  {\color[HTML]{00008B}\textbf{(-9.8)}}  & 57.4  {\color[HTML]{00008B}\textbf{(-10.5)}} \\ 
\textbf{SLW}  & 91.9 {\color[HTML]{00008B}\textbf{(-3.9)}}  & 97.0  {\color[HTML]{00008B}\textbf{(-1.1)}}  & 83.0  {\color[HTML]{00008B}\textbf{(-1.6)}}  & 66.7  {\color[HTML]{00008B}\textbf{(-1.1)}} \\ 
\textbf{WIF}  & 94.3  {\color[HTML]{00008B}\textbf{(-1.5)}}  & 97.7  {\color[HTML]{00008B}\textbf{(-0.5)}}  & 82.6  {\color[HTML]{00008B}\textbf{(-2.0)}}  & 65.6  {\color[HTML]{00008B}\textbf{(-2.2)}} \\ 
\textbf{SPP}  & 94.8  {\color[HTML]{00008B}\textbf{(-1.0)}}  & 97.7  {\color[HTML]{00008B}\textbf{(-0.5)}}  & 78.3  {\color[HTML]{00008B}\textbf{(-6.2)}}  & 65.5  {\color[HTML]{00008B}\textbf{(-2.4})} \\ \bottomrule 
\end{tabular}
\textbf{}}
\caption{Robustness test on Sentiment Analysis and CQA tasks. We apply four perturbations: \textbf{UPC} (upper casing), \textbf{SLW} (slang word), \textbf{WIF} (word inflection), and \textbf{SPP} (sentence paraphrasing).}
\label{tab:robustness}
\end{table}

\section{Discussion}
\label{sec:analysis}

\begin{table}[t]
\centering
\resizebox{0.75\linewidth}{!}{%
\begin{tabular}{llcccc}
\toprule
\multirow{2.3}{*}{\textbf{Datasets}} & \multicolumn{2}{c}{\textbf{QuAC Model}} & \multicolumn{2}{c}{\textbf{CANARD Model}} \\  \cmidrule(lr){2-3} \cmidrule(lr){4-5}
& \textbf{F1} & \textbf{EM} & \textbf{F1} & \textbf{EM} \\ \midrule
QuAC & 67.7 & 51.5 & 62.9 & 46.8  \\
CANARD & 65.1 & 49.9 & 63.3 & 46.9 \\ \bottomrule
\end{tabular}%
}
\caption{Results of the supervised learning approach. ``XX Model'' denotes the QA model trained on XX, and EM the percentage of the predictions the same as the gold.}
\vspace{-10pt}
\label{tab:canard_quac}
\end{table}

In this section, we provide an analysis to (i) raise a sensitivity problem of the QA model to explain the failure of RL and (ii) disclose that there is no justification for QR, even in the non-RL approaches.

\subsection{Sensitivity of the QA model}
It appears that the QA models are more sensitive to trivial changes than the reward models in other successful language generation tasks, and this could account for our lower performance on CQA. As can be seen from the examples in Table~\ref{tab:case_study}, a subtle alteration such as uppercasing or replacement with synonyms can significantly change F1 scores.

To quantify the sensitivity of the reward models, we compare model robustness between our QA models and sentiment analysis models that have been reported in~\citet{ziegler2019finetuning} to be effective for stylistic language generation.
We adopt publicly available models that are fine-tuned sentiment analysis datasets: BERT-based trained on Amazon polarity~\cite{mcauley2013amazon}\footnote{\url{https://huggingface.co/fabriceyhc/bert-base-uncased-amazon_polarity}} and RoBERTa-base trained on Yelp polarity~\cite{zhang2015yelp}\footnote{\url{https://huggingface.co/VictorSanh/roberta-base-finetuned-yelp-polarity}}.
To test the robustness of the models, we introduce small perturbations to the samples in the test set using the NL-Augmenter toolkit~\cite{dhole2021nlaugmenter}, and compare F1 scores on each task (experimental details in Appendix~\ref{app:robustness}).

Based on the robustness test given in Table~\ref{tab:robustness}, the QA models are shown to be significantly less robust against most perturbations compared to the sentiment analysis models.
It is conceivable that this sensitivity of the QA model leads to a sparse reward problem for the agent, which causes instability for the model learning the optimal policy. 
An important direction for future studies is to ease the sparse reward problem by, for example, enhancing the robustness of the QA models.


\subsection{Can QR Help in Non-RL Approaches?}
First, we evaluate with a simple supervised learning approach using rewrites provided by CANARD.
Extracting the QuAC samples that have a CANARD annotation, we (i) evaluate the CANARD annotations with the QA model trained on QuAC (the model used in the main experiments) and (ii) train another QA model with the CANARD annotations. Training is under the same conditions of the QA model initialization as in the main experiments.
As the results in Table~\ref{tab:canard_quac} show, we can hardly observe the effectiveness of the CANARD annotations. 
This supports the claim in~\citet{buck2018ask} that better rewrites in the human eye are not necessarily better for machines and implies the difficulty of exploiting QR for CQA. 

\begin{table}[t]
\centering
\resizebox{0.7\linewidth}{!}{%
\begin{tabular}{llcccc}
\toprule
\multirow{2.3}{*}{\textbf{Datasets}} & \multicolumn{2}{c}{\textbf{CoQA}} & \multicolumn{2}{c}{\textbf{QuAC}} \\  \cmidrule(lr){2-3} \cmidrule(lr){4-5}
& \textbf{F1} & \textbf{EM} & \textbf{F1} & \textbf{EM} \\ \midrule
end-to-end & \textbf{84.5} & \textbf{76.4} & \textbf{67.83} & 51.47 \\ \midrule
QReCC & 84.1 & 76.0 & \textbf{67.83} & 51.48  \\
CANARD & 83.7 & 75.8 & 67.81 & \textbf{51.50} \\ \bottomrule
\end{tabular}%
}
\caption{Results of the data augmentation approach. EM denotes the percentage of the predictions the same as the gold.}
\label{tab:data_aug}
\end{table}

Moreover, we explore a data-augmentation approach to integrate QR and CQA. First, we generate ten possible rewrites using top-$k$ sampling~\cite{zhang-etal-2021-trading} for all the questions of the CQA datasets. To guarantee the quality of the rewrites, we select the best F1 scoring ones from every ten candidates and use them to teach another QR model how to reformulate questions (experimental details in Appendix~\ref{app:augmentation}). As the results in Table~\ref{tab:data_aug} show, we consistently get worse scores compared to the end-to-end settings in CoQA, and almost the same scores for QuAC, not finding justification to apply QR in the manner of the data augmentation approach. 

\section{Conclusion}
In this paper, we explore the RL approach to verify the effectiveness of QR in CQA, and report that the RL approach is on par with simple end-to-end baselines.
We find the sensitivity of the QA models would disadvantage the RL training. Future work is needed to verify that QR is a promising mitigation method for CQA since even the non-RL approaches perform unsatisfactorily.



\bibliography{anthology,custom}
\bibliographystyle{acl_natbib}

\clearpage

\appendix

\pagenumbering{roman}
\counterwithin{table}{section}
\setcounter{table}{0}
\counterwithin{table}{section}
\setcounter{figure}{0}
\counterwithin{figure}{section}
\setcounter{footnote}{0}
\counterwithin{footnote}{section}

\section{Additional Details of Experiments}
\label{app:main_experiments}
Our implementation is based on~\citet{wolf-etal-2020-transformers}, and we plan to release our codes as well as trained models.
Before applying our reinforcement learning training, the QA and the QR models are initialized with the best QA and QR models. The QA models are trained on CoQA and QuAC datasets, and model selection is based on their F1 score. The QR models are trained on QReCC and CANARD, and model selection is based on the BLEU~\cite{papineni2002bleu}\footnote{\url{https://github.com/moses-smt/mosesdecoder/blob/master/scripts/generic/multi-bleu-detok.perl}} score for CANARD and ROUGE-1R~\cite{lin-2004-rouge} for QReCC, respectively. We report the other hyperparameters in Table~\ref{tab:QA_model_hyperparams}
and Table~\ref{tab:QR_model_hyperparams}. We use Adam optimizer~\cite{kingma2015adam} for all the training.

The hyperparameters used for the PPO training are reported in Table~\ref{tab:all_hyperparams}.
For rewrites generation with the QR model, we use beam search with beam width of 5, preventing generation repetition \cite{DBLP:journals/corr/abs-1909-05858} with using repetition penalty of 1.1, and set the maximum input sequence length to 512. 
We then run the PPO with value function coefficient of 1.0, while ensuring the sequence length of question rewriting model input to be 150 tokens maximum and the generations length to be 50 tokens at maximum. To ensure that the learned policy does not deviate too much, we apply an additional reward signal, adaptive KL factor $\beta$ according to the magnitude of the KL-penalty with a KL-coefficient $K_\beta = 0.1$ following~\citet{ziegler2019finetuning}.

\begin{table}[!t]
\centering
\resizebox{0.9\linewidth}{!}{%
\begin{tabular}{lcc}
\toprule
\textbf{Hyperparameter settings} & \textbf{CoQA} & \textbf{QuAC} \\ \midrule
Model architecture & \multicolumn{2}{c}{RoBERTa-base} \\
Model size & \multicolumn{2}{c}{125M parameters} \\
Optimizer & \multicolumn{2}{c}{Adam} \\
Learning rate & \multicolumn{2}{c}{$3\mathrm{e}{-5}$} \\
Warmup steps & 1000 & 0 \\
Weight decay & 0.01 & 0.01 \\
Gradient accumulation steps & 10 & 20 \\
Early stopping patience & 3 & 4 \\
Batch size & \multicolumn{2}{c}{6} \\
Maximum epoch & \multicolumn{2}{c}{10} \\
Document stride & \multicolumn{2}{c}{128} \\
Maximum sequence length & \multicolumn{2}{c}{512} \\
Maximum answer length & \multicolumn{2}{c}{50} \\
\bottomrule
\end{tabular}%
}
\caption{Hyperparameters for initialization of QA models.}
\label{tab:QA_model_hyperparams}
\end{table}

\begin{table}[!t]
\centering
\resizebox{0.9\linewidth}{!}{%
\begin{tabular}{lcc}
\toprule
\textbf{Hyperparameter settings} & \textbf{Value} \\ \midrule
Model architecture & GPT-2 base \\
Model size & 117M parameters \\
Optimizer & Adam \\
Learning rate & $3\mathrm{e}{-5}$ \\
Warmup steps & 500 \\
Batch size & 8 \\
Gradient accumulation steps & 8 \\
Early stopping patience & 3 \\
Maximum epoch & 20 \\
History length (utterances) & 3 \\
Maximum sequence length & 256 \\
\bottomrule
\end{tabular}%
}
\caption{Hyperparameters for initialization of QR models both in QRECC and CANARD dataset training.}
\label{tab:QR_model_hyperparams}
\end{table}

\begin{table}[!t]
\centering
\resizebox{0.9\linewidth}{!}{%
\begin{tabular}{lcccc}
\toprule
\textbf{Hyperparameter settings} & \textbf{Value}\\ \midrule
\textbf{Training settings} &  \\
Optimizer & Adam \\
Learning rate & 1E-5, 1E-6, (1E-7), 1E-8, 1E-9 \\
Batch size & 8 (CoQA), 16 (QuAC) \\
Early stopping patience & 3 \\
Max epoch & 6 \\
QA history length & 3 \\
Max query rewrite length & 100 \\
QR max sequence length & 150 \\
Max sequence length & 512 \\
Max question length & 128 \\
Pad to maximum length & TRUE \\
Document stride & 128 \\
N best answers to generate & 20 \\
Max answer length & 5 \\
\midrule
\textbf{PPO settings} & \\
Maximum PPO epoch & 4\\
KL coefficient init & 0.2\\
Target & 6\\
Horizon & 10000\\
gamma & 1\\
lambda & 0.95\\
cliprange & 0.2\\
vf\_coef & 0.5\\
ce\_coef & 1\\
\bottomrule
\end{tabular}%
}
\caption{Hyperparameters for PPO training and inference (Best parameter is in bracket).}
\label{tab:all_hyperparams}
\end{table}

\section{Details of Robustness Experiment}
\label{app:robustness}
We perform robustness testing by introducing a minor perturbation which minimally change the semantic of the sentence. We utilize NL-Augmenter toolkit~\cite{dhole2021nlaugmenter}~\footnote{\url{https://github.com/GEM-benchmark/NL-Augmenter}} to generate 5 different type of perturbations, i.e., random upper casing~\cite{wei2019eda}~\footnote{\url{https://github.com/GEM-benchmark/NL-Augmenter/tree/main/transformations/random_upper_transformation}}, contraction expansion~\cite{ribeiro2020beyond}~\footnote{\url{https://github.com/GEM-benchmark/NL-Augmenter/tree/main/transformations/contraction_expansions}}, word inflection variation~\cite{tan2020morphin}~\footnote{\url{https://github.com/GEM-benchmark/NL-Augmenter/tree/main/transformations/english_inflectional_variation}}, slang word~\footnote{\url{https://github.com/GEM-benchmark/NL-Augmenter/tree/main/transformations/slangificator}}, and sentence paraphrasing using yoda transformation~\footnote{\url{https://github.com/GEM-benchmark/NL-Augmenter/tree/main/transformations/yoda_transform}} perturbations. For random upper casing, we apply a probability of 10\% to randomly upper case letters. For contraction expansion, inflection variation, and slang word variation perturbations, we apply them without randomness by replacing all occurrences of words that fulfill the corresponding rules of each perturbation so as to increase the difference to the original sentence. For sentence paraphrasing, we apply a rule-based yoda paraphrasing which reconstruct the sentence into its ``XSV'' syntax format, where the `S' stands for subject, the `V' stands for verb, and the `X' being a stand-in for whatever chunk of the sentence goes after the verb from the original ``SVX'' syntax format. Table~\ref{tab:app_robustness} shows the BLEU score~\cite{papineni2002bleu} and the Levensthein distance~\cite{miller2002levensthein} from each perturbation compared to the original sentence in each dataset. We do not report contraction expansion perturbation due to the minute differences with the original sentence.

\begin{table}[!t]
\centering
\resizebox{\linewidth}{!}{%
\begin{tabular}{lcccccccc}
\toprule
\multirow{2}{*}{\textbf{Perturb}} & \multicolumn{2}{c}{\textbf{Amazon}} & \multicolumn{2}{c}{\textbf{Yelp}} & \multicolumn{2}{c}{\textbf{CoQA}} & \multicolumn{2}{c}{\textbf{QuAC}} \\ \cmidrule(lr){2-9} 
 & \textbf{LD} & \textbf{BLEU} & \textbf{LD} & \textbf{BLEU} & \textbf{LD} & \textbf{BLEU} & \textbf{LD} & \textbf{BLEU} \\
\midrule
\textbf{UPC}  & 1.0\% & 94.0 & 0.2\% & 98.1 & 1.2\% & 93.6 & 0.8\% & 95.6 \\
\textbf{WCE}  & 7.3\% & 43.1 & 7.4\% & 34.4 & 6.8\% & 36.7 & 7.1\% & 33.8 \\
\textbf{SLW}  & 5.0\% & 89.9 & 5.1\% & 86.8 & 3.5\% & 91.3 & 1.5\% & 95.7 \\
\textbf{WIF}  & 8.8\% & 52.7 & 7.9\% & 54.6 & 10.4\% & 44.0 & 9.1\% & 47.8 \\
\textbf{SPP}  & 31.1\% & 57.3 & 32.7\% & 68.5 & 53.4\% & 34.6 & 46.4\% & 39.9 \\ \bottomrule
\end{tabular}
}
\caption{Perturbation statistics on  Amazon, Yelp, CoQA, and QuAC tasks. \textbf{LD} denotes the Levensthein distance divided by the number of characters in the original dataset. \textbf{BLEU} denotes BLEU score. We apply five perturbations: \textbf{UPC} (upper casing), \textbf{WCE} (word contraction expansion), \textbf{SLW} (slang word), \textbf{WIF} (word inflection), and \textbf{SPP} (sentence paraphrasing).}
\label{tab:app_robustness}
\end{table}

\section{Details of Data Augmentation Approach}
\label{app:augmentation}
\begin{table*}[!t]
\centering
\resizebox{0.9\linewidth}{!}{
\begin{tabular}{lccccccccc}
    \toprule
    \multicolumn{1}{l}{\multirow{2}{*}{\textbf{Settings}}} & \multicolumn{6}{c}{\textbf{CoQA}} & \multicolumn{3}{c}{\textbf{QuAC}} \\ \cmidrule(lr){2-7} \cmidrule(lr){8-10}
    \multicolumn{1}{l}{} & \textbf{Overall F1} & \textbf{Child.} & \textbf{Liter.} & \textbf{M\&H} & \textbf{News} & \textbf{Wiki.} & \textbf{F1} & \textbf{HEQ-Q} & \textbf{HEQ-D} \\ \midrule
    end-to-end & \textbf{84.5} & \textbf{84.4} & \textbf{82.4} & \textbf{82.9} & \textbf{86.0} & \textbf{86.9} & \textbf{67.8} & \textbf{63.5} & 7.9 \\ 
    QReCC-augment & 84.1 & 84.1 & 81.9 & 82.4 & 85.4 & 86.7 &\textbf{67.8} & 63.4  & \textbf{8.1} \\ 
    CANARD-augment & 83.7 & 83.8 & 81.6 & 81.9 & 85.1 & 86.4 & \textbf{67.8} & \textbf{63.5} & 7.8 \\ 
    \bottomrule
    \end{tabular}
}
\caption{Evaluation results of the data augmentation methods. Our data augmentation approach scores constantly worse than the end-to-end setting for CoQA, and almost the same for QuAC.}
\label{tab:results_aug}
\end{table*}

We investigate a data augmentation approach. While we adopt the question rewrites obtained by the QR model to fine-tune the QA model regardless of their quality, we filter rewrites based on their F1 scores.
We first generate 10 possible rewrites using $k=20$ and $p=0.95$ with top-$k$ sampling~\cite{zhang-etal-2021-trading} for all the questions in the training and validation set of CoQA and QuAC with the QR model pretrained either on QReCC or CANARD. To guarantee the quality of the rewrites, we select the best scoring one out of 10 candidates on the corresponding CQA or use the original question if the original question gets a better F1 score.
Then, we finetune another QR model using the newly annotated question rewrites pairs $\{X_t, \acute{Q}_t\}$ so that the model learns how to reformulate questions. Here, following our main experiments, we use pretrained GPT-2 as initialization for finetuning. We use Adam optimizer~\cite{kingma2015adam} with the learning rate of $3\mathrm{e}{-5}$, and the other hyperparameters follow Table~\ref{tab:QR_model_hyperparams}.

We report the evaluation results on CoQA and QuAC in Table~\ref{tab:results_aug}. We cannot find any justification to apply the data augmentation approach from our experiments. We constantly get worse scores compared to the end-to-end settings in CoQA, and almost the same for QuAC. 
We could not observe any meaningful improvement, except that it performs better than the pipeline settings (see Table 1 for comparison).

\section{Additional Analysis: Gradient Saliency Analysis}
Inspired by~\citet{mudrakarta-etal-2018-model}, we construct a saliency map for each sample in CoQA to observe which part of the inputs counts for the predictions using Integrated Gradient analysis \citep{kokhlikyan2020captum, sundararajan2017axiomatic}.
Based on our QA model implementation, we construct a saliency map for the predictions on a start index, end index, and on a label to determine whether it is a yes/no/unknown question. 

\begin{figure*}[t]
    \centering
    \begin{subfigure}
         \centering
         \includegraphics[width=0.97\linewidth]{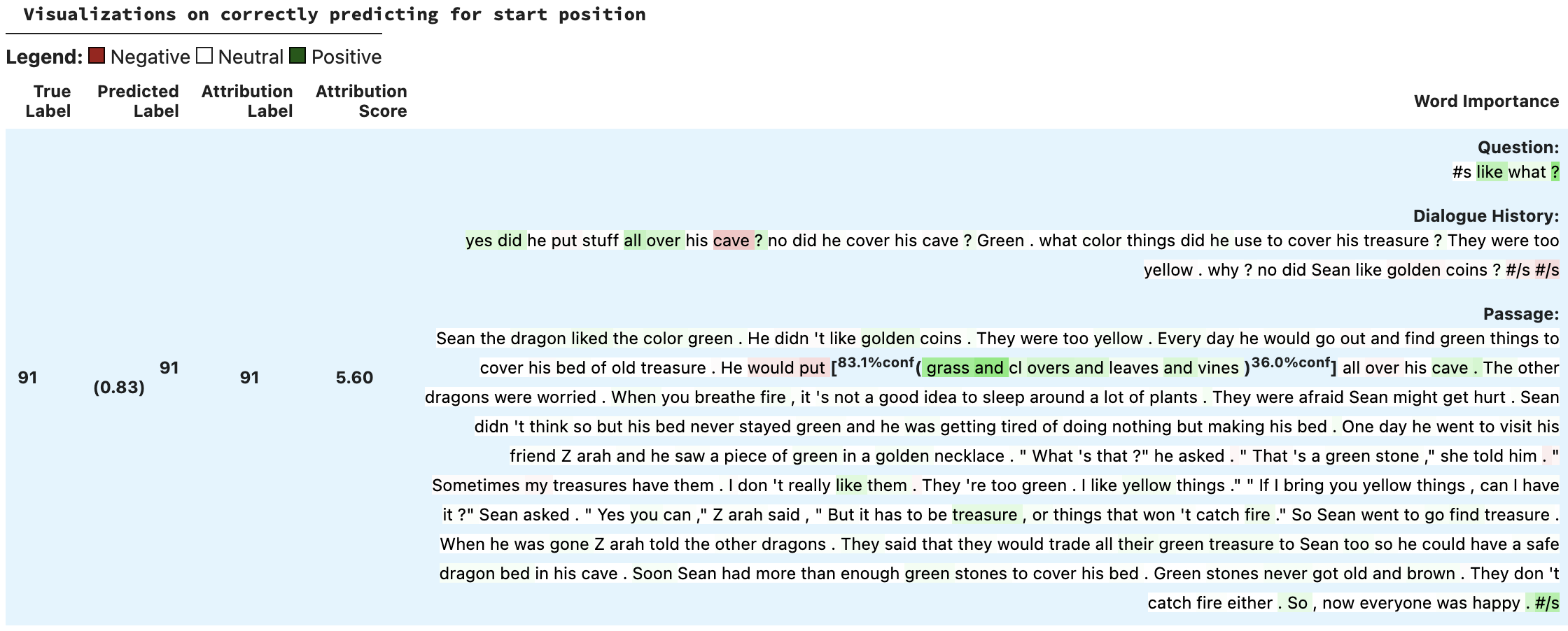}
         \caption{Visualization of Integrated Gradient saliency analysis on our RL approach that predicts the start index correctly. Round brackets indicate the QA model predictions of the start and end index, and square brackets the gold start and end index.}
         \label{fig:start_grad_vis}
     \end{subfigure}
     \begin{subfigure}
         \centering
         \includegraphics[width=0.97\linewidth]{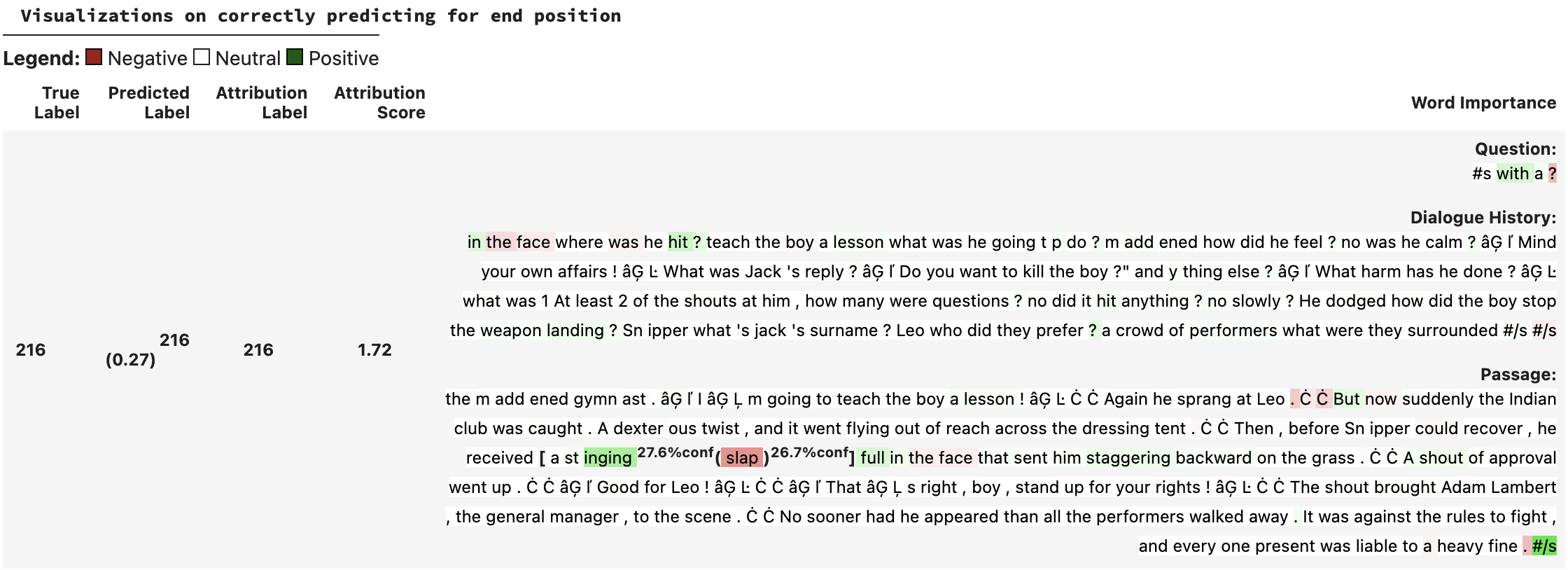}
         \caption{Visualization of Integrated Gradient saliency analysis on our RL approach that predicts the end index correctly. Round brackets indicate the QA model predictions of the start and end index, and square brackets the gold start and end index.}
         \label{fig:end_grad_vis}
     \end{subfigure}
     \begin{subfigure}
         \centering
         \includegraphics[width=0.97\linewidth]{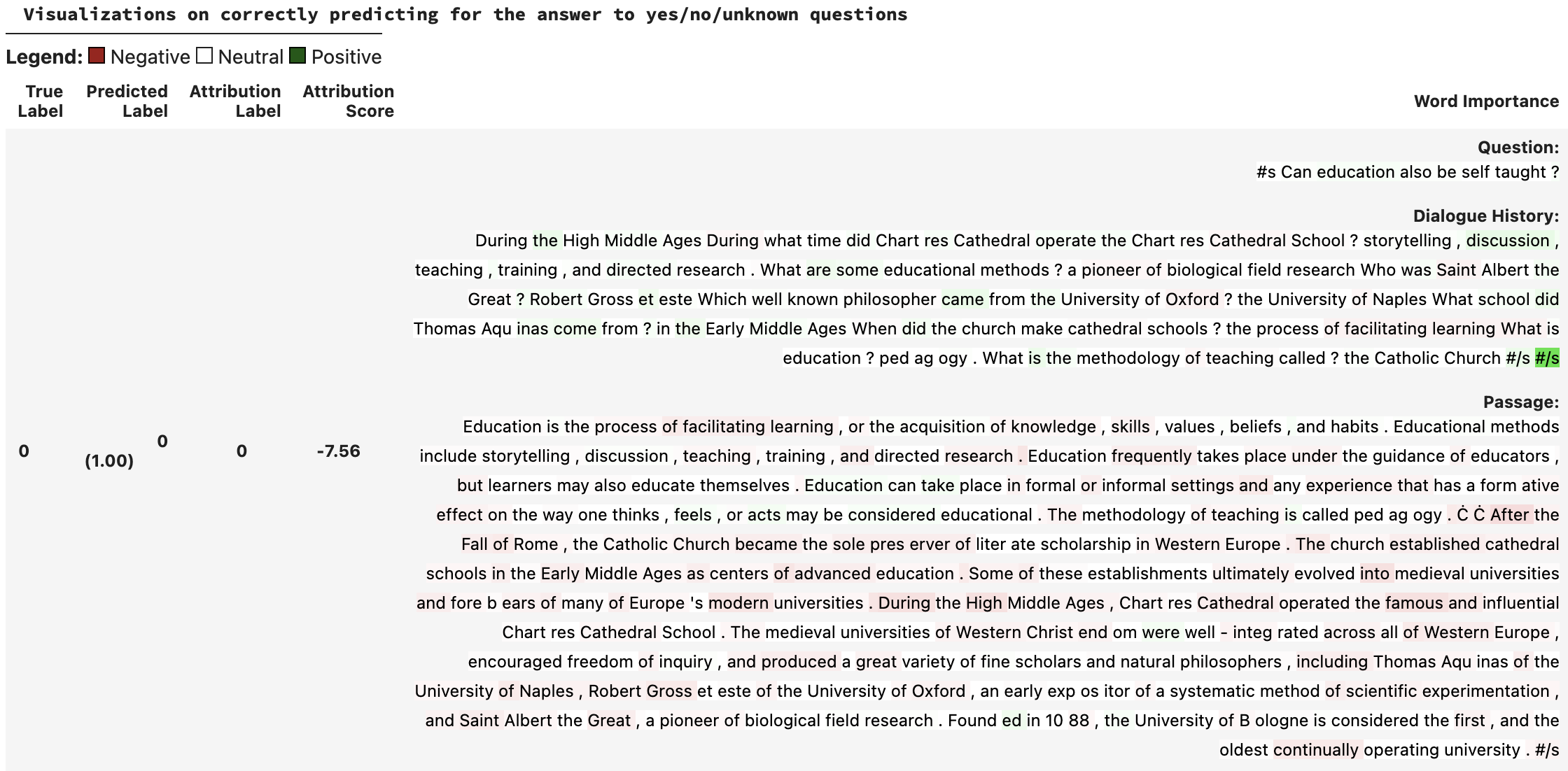}
         \caption{Visualization of Integrated Gradient saliency analysis on our RL approach that correctly the answers yes/no/unknown question.}
         \label{fig:ynu_grad_vis}
     \end{subfigure}
    \label{fig:correlation}
\end{figure*}

Examples of the saliency map are depicted in Figure \ref{fig:start_grad_vis}, \ref{fig:end_grad_vis}, and \ref{fig:ynu_grad_vis} respectively. For the sake of better interpretability, we compute the distribution of tokens that affects the QA model predictions either positively or negatively. We first count each tokens' positive ($>0$) and negative ($<0$) attributions, and then normalized with each input lengths. The normalized value then taken as a distribution over questions, dialogue histories, and contexts. 
Results are reported in Table~\ref{tab:rewritten_non-zero_attributions_distribution_normalized} - \ref{tab:original_significant_attributions_distribution}. Note that the total per column is not always 100\% due to rounding.
From the comparison between Table~\ref{tab:rewritten_non-zero_attributions_distribution_normalized} and Table~\ref{tab:original_non-zero_attributions_distribution_normalized} or between Table~\ref{tab:rewritten_significant_attributions_distribution} and Table~\ref{tab:original_significant_attributions_distribution}, we only observe marginal differences between the distributions. It is linked to the issue we report in the main paper that the QR models learned to copy the original questions.

From the comparison between Table~\ref{tab:rewritten_non-zero_attributions_distribution_normalized} and Table~\ref{tab:rewritten_significant_attributions_distribution} or between Table~\ref{tab:original_non-zero_attributions_distribution_normalized} and Table~\ref{tab:original_significant_attributions_distribution}, we can see that the dialogue histories have more effects on the model predictions than questions, however, the questions give more significant contributions to them. Throughout the gradient saliency analysis, we cannot find any justification to apply QR for CQA, since our results could be interpreted as the QA model can leverage the dialogue history without QR.

\begin{table}[!t]
\centering
\resizebox{\linewidth}{!}{%
\begin{tabular}{lcccccc}
\toprule
\multirow{3}{*}{\textbf{Input}} & \multicolumn{6}{c}{\textbf{Distribution of Contributing Tokens}} \\ \cmidrule(lr){2-7}
 & \multicolumn{2}{c}{\textbf{start index}} & \multicolumn{2}{c}{\textbf{end index}} & \multicolumn{2}{c}{\textbf{yes/ no/ unknown}} \\ \cmidrule(lr){2-3} \cmidrule(lr){4-5} \cmidrule(lr){6-7}
 & \textbf{\textgreater 0} & \textbf{\textless 0} & \textbf{\textgreater 0} & \textbf{\textless 0} & \textbf{\textgreater 0} & \textbf{\textless 0} \\ \midrule
\multicolumn{1}{r}{$\acute{Q}_t$} & 32.7\% & 25.9\% & 29.9\% & 30.0\% & 3.5\% & 0.8\% \\
\multicolumn{1}{r}{$\mathcal{D}_{t-1}$} & 33.2\% & 37.5\% & 33.5\% & 37.5\% & 28.6\% & 14.0\% \\
\multicolumn{1}{r}{$Y_t$} & 33.9\% & 36.5\% & 36.5\% & 32.4\% & 67.7\% & 85.1\%
\\ \bottomrule 
\end{tabular}}
\caption{Results of Integrated Gradient Analysis on the CoQA inputs using the rewritten questions.}
\label{tab:rewritten_non-zero_attributions_distribution_normalized}
\end{table}

\begin{table}[!t]
\centering
\resizebox{\linewidth}{!}{%
\begin{tabular}{lcccccc}
\toprule
\multirow{3}{*}{\textbf{Input}} & \multicolumn{6}{c}{\textbf{Distribution of Contributing Tokens}} \\ \cmidrule(lr){2-7}
 & \multicolumn{2}{c}{\textbf{start index}} & \multicolumn{2}{c}{\textbf{end index}} & \multicolumn{2}{c}{\textbf{yes/ no/ unknown}} \\ \cmidrule(lr){2-3} \cmidrule(lr){4-5} \cmidrule(lr){6-7}
 & \textbf{\textgreater 0.5} & \textbf{\textless{}= -0.5} & \textbf{\textgreater 0.5} & \textbf{\textless{}= -0.5} & \textbf{\textgreater 0.5} & \textbf{\textless{}= -0.5} \\ \midrule
\multicolumn{1}{r}{$\acute{Q}_t$} & 3.9\% & 0.9\% & 8.4\% & 1.0\% & 0.0\% & 0.0\% \\
\multicolumn{1}{r}{$\mathcal{D}_t$} & 1.4\% & 1.7\% & 1.9\% & 1.0\% & 100.0\% & 100.0\% \\
\multicolumn{1}{r}{$Y_t$} & 94.6\% & 97.2\% & 89.5\% & 97.9\% & 0.0\% & 0.0\%\\ \bottomrule 
\end{tabular}}
\caption{Results of Integrated Gradient Analysis on the CoQA inputs using the rewritten questions for significantly contributing tokens.}
\label{tab:rewritten_significant_attributions_distribution}
\end{table}

\begin{table}[!t]
\centering
\resizebox{\linewidth}{!}{%
\begin{tabular}{lcccccc}
\toprule
\multirow{3}{*}{\textbf{Input}} & \multicolumn{6}{c}{\textbf{Distribution of Significantly Contributing Tokens}} \\ \cmidrule(lr){2-7}
 & \multicolumn{2}{c}{\textbf{start index}} & \multicolumn{2}{c}{\textbf{end index}} & \multicolumn{2}{c}{\textbf{yes/ no/ unknown}} \\ \cmidrule(lr){2-3} \cmidrule(lr){4-5} \cmidrule(lr){6-7}
 & \textbf{\textgreater 0} & \textbf{\textless 0} & \textbf{\textgreater 0} & \textbf{\textless 0} & \textbf{\textgreater 0} & \textbf{\textless 0} \\ \midrule
\multicolumn{1}{r}{$Q_t$} & 32.8\% & 25.6\% & 29.8\% & 29.8\% & 3.5\% & 0.8\% \\
\multicolumn{1}{r}{$\mathcal{D}_t$} & 33.2\% & 37.7\% & 33.5\% & 37.6\% & 28.7\% & 14.0\% \\
\multicolumn{1}{r}{$Y_t$} & 33.9\% & 36.6\% & 36.5\% & 32.5\% & 67.6\% & 85.1\%
\\ \bottomrule 
\end{tabular}}
\caption{Results of Integrated Gradient Analysis on the CoQA inputs using the original questions.}
\label{tab:original_non-zero_attributions_distribution_normalized}
\end{table}

\begin{table}[!t]
\centering
\resizebox{\linewidth}{!}{%
\begin{tabular}{lcccccc}
\toprule
\multirow{3}{*}{\textbf{Input}} & \multicolumn{6}{c}{\textbf{Distribution of Significantly Contributing Tokens}} \\ \cmidrule(lr){2-7}
 & \multicolumn{2}{c}{\textbf{start index}} & \multicolumn{2}{c}{\textbf{end index}} & \multicolumn{2}{c}{\textbf{yes/ no/ unknown}} \\ \cmidrule(lr){2-3} \cmidrule(lr){4-5} \cmidrule(lr){6-7}
 & \textbf{\textgreater 0.5} & \textbf{\textless{}= -0.5} & \textbf{\textgreater 0.5} & \textbf{\textless{}= -0.5} & \textbf{\textgreater 0.5} & \textbf{\textless{}= -0.5} \\ \midrule
\multicolumn{1}{r}{$Q_t$} & 3.9\% & 1.0\% & 8.2\% & 1.0\% & 0.0\% & 0.0\% \\
\multicolumn{1}{r}{$\mathcal{D}_t$} & 1.6\% & 2.0\% & 2.0\% & 1.0\% & 100.0\% & 100.0\% \\
\multicolumn{1}{r}{$Y_t$} & 94.4\% & 96.9\% & 89.6\% & 97.9\% & 0.0\% & 0.0\% 
\\ \bottomrule 
\end{tabular}}
\caption{Results of Integrated Gradient Analysis on the CoQA inputs using the original questions for significantly contributing tokens.}
\label{tab:original_significant_attributions_distribution}
\end{table}

\clearpage

\end{document}